\pdfoutput=1

\documentclass[11pt]{article}

\usepackage{acl}

\usepackage{times}
\usepackage{latexsym}
\usepackage{graphicx}
\usepackage{comment}
\usepackage[T1]{fontenc}

\usepackage[utf8]{inputenc}

\usepackage{microtype}
\usepackage{inconsolata}

\usepackage{array}
\usepackage{ragged2e}
\newcolumntype{L}[1]{>{\raggedright\let\newline\\\arraybackslash\hspace{0pt}}p{#1}}
\newcolumntype{M}[1]{>{\centering\arraybackslash}m{#1}}
\usepackage{paralist}

\newcommand{\framework}{$\texttt{ECBD}$}

\title{\framework{}: Evidence-Centered Benchmark Design for NLP}

\author{\textbf{Yu Lu Liu}$^{1,2}$~~~~\textbf{Su Lin Blodgett}$^{3}$~~~~\textbf{Jackie Chi Kit Cheung}$^{1,2,4}$\\\textbf{Q. Vera Liao}$^{3}$~~~~\textbf{Alexandra Olteanu}$^{3}$~~~~\textbf{Ziang Xiao}$^{3,5}$ \\
    $^1$Mila -- Quebec Artificial Intelligence Institute \quad
    $^2$McGill University \\
    $^3$Microsoft Research, Montr\'{e}al, Canada \quad
    $^4$Canada CIFAR AI Chair \quad
    $^5$Johns Hopkins University\\
    \texttt{yu.l.liu@mail.mcgill.ca}\quad \texttt{jackie.cheung@mcgill.ca}\\
    \texttt{\{sulin.blodgett,veraliao,alexandra.olteanu\}@microsoft.com}\\
    \texttt{ziang.xiao@jhu.edu}
}

\begin{document}
\maketitle
\begin{abstract}
Benchmarking is seen as critical to assessing progress in NLP. 
However, creating a benchmark involves many design decisions (e.g., which datasets to include, which metrics to use) that often rely on tacit, untested assumptions about what the benchmark is intended to measure or is actually measuring.
There is currently no principled way of analyzing these decisions and how they impact the validity of the benchmark's measurements.
To address this gap, we draw on evidence-centered design in educational assessments and propose Evidence-Centered Benchmark Design (\framework{}), a framework which formalizes the benchmark design process into five modules. 
\framework{} specifies the role each module plays in helping practitioners collect evidence about capabilities of interest.
Specifically, each module requires benchmark designers to describe, justify, and support benchmark design choices---e.g., clearly specifying the capabilities the benchmark aims to measure or how evidence about those capabilities is collected from model responses.
To demonstrate the use of \framework{}, we conduct case studies with three benchmarks: BoolQ, SuperGLUE, and HELM. Our analysis reveals common trends in benchmark design and documentation that could threaten the validity of benchmarks' measurements.
\end{abstract}

\section{Introduction}
While benchmarking has long been seen as critical to gauging progress in natural language processing (NLP) and guiding model selection for downstream applications, assessing the quality of a benchmark remains a persistent challenge.
Do benchmark measurements---most often in the form of numerical scores---provide meaningful insights about the evaluated models and their capabilities? How valid are these measurements? For what purposes are they useful?
The field of NLP lacks a systematic way of reflecting on these important questions. 

\begin{figure}[t]
    \centering
    \includegraphics[trim= 0 2.5 0 0, clip,scale=0.7]{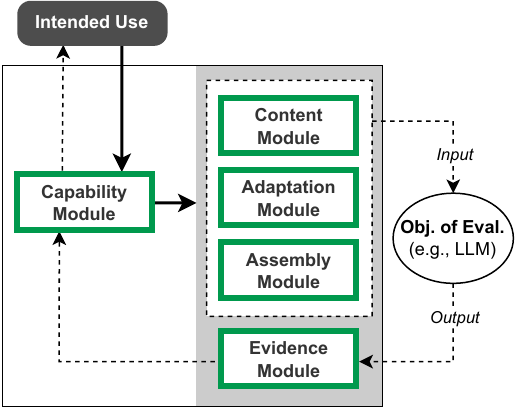}
    \caption{Simplified schema of the Evidence-Centered Benchmark Design (\framework{}) framework. Solid line arrows indicate the process of designing a benchmark (e.g., designers should decide on the intended uses of the benchmark before deciding what capabilities are of interest). The dotted line arrows indicate the process wherein the benchmark gathers necessary evidence.}
    \vspace{-12pt}
    \label{fig:ecbd_small_schema}
\end{figure}

At the same time, as NLP models are increasingly believed to be more performant and to exhibit a wider range of capabilities, 
evaluation in NLP has shifted from measuring model performance on a specific dataset for a single task to using large benchmarks that cover multiple tasks (e.g., GLUE \cite{wang-etal-2018-glue}, SuperGLUE \cite{superglue}, BIG-Bench \cite{srivastava2022imitation}, HELM \cite{liang2022holistic}). These benchmarks are increasingly larger and more ambitious (e.g., HELM aims to ``assess language models in their totality''), covering ever-growing numbers of tasks, datasets, and metrics aimed at measuring an increasing number of capabilities---i.e., abilities or behaviors that researchers believe the models exhibit or might exhibit. This trend further increases the complexity of assessing benchmark quality.

Such issues with assessing measurement quality do not only concern NLP practitioners. Researchers and practitioners in educational testing often face similar questions: do students' exam results provide meaningful insights about their ability in, for example, reading comprehension? Can these results be used to determine whether a student needs remedial classes?

In this work, we take inspiration from the Evidence-Centered Design (ECD) framework in educational testing \cite{mislevy_structure_ECD}. ECD views testing as the process of gathering evidence about students' abilities, and provides guidance on the creation, documentation, and validation of educational tests. We draw an analogy between educational tests and NLP benchmarks and propose the Evidence-Centered Benchmark Design (\framework{}) framework, in which we view benchmarking as the process of gathering evidence from objects of evaluation (e.g., language models) about whether or to what degree they have some capabilities of interest.

\framework{} unpacks and formalizes benchmark design decisions into five modules, each having a specific role in supporting the process of collecting necessary 
evidence (see Figure~\ref{fig:ecbd_small_schema}). For each module, we provide guiding questions that help benchmark creators document, justify, and validate their design choices. These same questions can also guide benchmark users in analyzing and documenting existing benchmarks to better understand their limitations and how to appropriately use them: what are the design decisions shaping the benchmark? Why did its creators make these decisions? And what evidence do they provide to support their decisions? 
By doing so, \framework{} also supports added transparency about what benchmarks measure, and when and how they can be used. In turn, this transparency can also help our community contest and mitigate issues with current benchmarks.

To illustrate \framework{}'s usage for benchmark analysis, we organize these questions into a worksheet (Appendix~\ref{appendix: worksheet_template}) and apply it to three different benchmarks: BoolQ \cite{clark-etal-2019-boolq}, SuperGLUE \cite{superglue}, and HELM \cite{liang2022holistic}.\footnote{Completed worksheets at \url{https://github.com/isle-dev/ECBD}} Through this exercise, we uncover common issues, such as poor conceptualization of capabilities, that threaten the validity of these benchmarks' measurements. In general, we find that these benchmarks lack justification and validation, as well as appropriate documentation of the many of their design choices made during their creation.\looseness=-1 

\section{Background \& Related Work}
\paragraph{Benchmarking in NLP}
At a time when most NLP models were built for a single specific task, \citet{wang-etal-2018-glue} introduced the General Language Understanding Evaluation (GLUE) benchmark with the goal of helping the research community develop models with better \textit{general} language understanding abilities. It includes a collection of nine English language understanding tasks, covering question answering, sentiment analysis, and textual entailment. 
Soon after, prompted by the belief that many models were already surpassing the performance of non-expert humans on GLUE, \citet{superglue} proposed SuperGLUE. 

This trend of evaluating models across an increasing number of datasets and tasks continues, with recent benchmarks such as GEM \cite{GEM_v1, GEM_v2}, covering 40 language generation tasks. Collaborative and evolving benchmarks such as BIG-Bench \cite{srivastava2022imitation}, now counting more than 200 tasks in its repository,\footnote{\url{https://github.com/google/BIG-bench}} call on the research community to contribute new tasks. 

We designed \framework{} to encourage more critical analyses of these increasingly complex benchmarks and a deeper reflection on how benchmark design decisions might affect the validity of benchmark measurements. 

\paragraph{Critiques and Meta-Analyses}
Prior work has also surveyed and critiqued both NLP and machine learning (ML) evaluation more generally. \citet{bowman-dahl-2021-will} outline a list of criteria they argue that useful benchmarks for natural language understanding (NLU) should meet, including validity. Similarly, \citet{ML_that_matters} highlights the disconnect between benchmark results and real world impacts---e.g., does a given increase on the benchmark actually lead to positive impact in the tested domain of application?---while \citet{liao2023rethinking} argue for centering language models' evaluation on how models will be used in practice. Analyses of benchmarks in NLP evaluation have raised concerns about annotation artifacts \cite{gururangan-etal-2018-annotation}, threats to validity \cite{blodgett-etal-2021-stereotyping}, lack of justification surrounding design choices \cite{goldfarb-tarrant-etal-2023-prompt}, inconsistent results from benchmarks aimed at measuring similar things \cite{akyurek-etal-2022-challenges}, and benchmarks' lack of robustness \cite{alzahrani2024when}.\looseness=-1

\paragraph{NLP and ML Documentation}
Various documentation guidelines have been proposed for NLP and ML datasets, models, and systems~\cite{bender-friedman-2018-data,arnold2019factsheets,mitchell2019model,datasheets,bhardwaj2024machine}. Datasheets for Datasets \cite{datasheets} provides a standardized process for documenting ML datasets, formulated as a list of questions (e.g., ``Does the dataset contain data that might be considered confidential?''). In NLP, Data Statements for NLP \cite{bender-friedman-2018-data} contains guidelines more specific to speech and text data, asking practitioners to document details about how data is curated, such as the demographics of the speakers included. Similarly, Model Cards \cite{mitchell2019model} and FactSheets~\cite{arnold2019factsheets} have been proposed to support better model and AI service documentation. 

Our work contributes to these efforts by providing a set of guidelines for documenting NLP benchmark design choices. Our focus, however, is on design choices that affect how benchmarks are used to gather the necessary evidence about whether, or to what degree, an evaluated model has some capabilities of interest. 
Our framework also guides the process of gathering the required validity evidence for benchmark measurements.

\paragraph{Measurement Theory}
In the social sciences, hypothesized theoretical entities known as \textbf{constructs} (e.g., a person's creativity, attitude towards a social issue) cannot be directly measured \citep{measurement_fairness}. Instead, the measurement is indirect, relying on samples of observable behaviors obtained through \textbf{tests}. Measurement theory is the study of test development, aiming to minimize measurement error so to produce the best measurements of the desired constructs \citep{measurement_theory_2018}. Educational testing is rooted in measurement theory, aiming to produce the best measurements of students' abilities.\looseness=-1

The quality of tests depends on their \textbf{validity}, which refers to ``the degree to which evidence and theory support the interpretations of test scores for proposed uses of tests'' \citep{standards}. \citet{measurement_theory_2018} argues that it is the most important quality of a test as it concerns the fundamental issue of what measurement instruments (i.e., tests) are really measuring.

These concepts are relevant to NLP, as many desirable model capabilities (e.g., language understanding) cannot be directly measured; they are unobservable constructs, and NLP benchmarks can be seen as tests that use observable model behaviors (e.g., LM-generated text) to measure these constructs \citep{xiao-etal-2023-evaluating-evaluation}. 
This raises key concerns about the validity of the performance measurements obtained with various NLP benchmarks~\cite{blodgett-etal-2021-stereotyping,bowman-dahl-2021-will,fleisig-etal-2023-fairprism}.  

\paragraph {Evidence-Centered Design (ECD) in Education}
is a framework introduced in the field of education with the goal of guiding the design, evaluation, and interpretation of educational tests~\cite{mislevy_structure_ECD}. Our main source of inspiration to create Evidence-Centered Benchmark Design (\framework{}) comes from the conceptual assessment framework (CAF), a vital component of ECD consisting of five models:
\begin{compactenum}[i)]
    \item \textbf{Student model:} specifies the constructs that characterize the students and that the test aims to measure. This model connects the test to its intended uses (e.g., if a test is to determine whether students need remedial language classes, should their reading comprehension skill be measured?).
    \item \textbf{Task model:} builds a pool of tasks (i.e., test items) that draw out responses from students. Since the test relies on these responses to measure the constructs of interest, the tasks should elicit evidence about those constructs. 
    \item \textbf{Presentation model:} specifies how a given test item or task is presented to students (e.g., font size, instructions given by teachers). The goal is to avoid introducing measurement error (e.g., due to differences in the readability of the test using inconsistent font sizes). 
    \item \textbf{Assembly model:} specifies how tasks are selected from the available pool to be presented to students (e.g., when there are 100 exam questions but students can only answer 20, how should the test select these 20 questions?). This model also specifies and helps assess 
    the amount of evidence that will be collected (e.g., are the selected 20 questions sufficient to measure reading comprehension?)  
    \item \textbf{Evidence model:} specifies how to measure constructs specified in the student model by observing students' performance on the presented test items. It consists of two components: one specifies item-level scoring (i.e., extracting evidence from students' performance on a single test item) and the other specifies test-level scoring (i.e., accumulating extracted evidence across all presented test items). 
\end{compactenum} 

In summary, each CAF model has specific roles to fulfill, and together they roadmap the process of educational testing. We adapt CAF models to NLP benchmarking, proposing a framework for benchmark design that similarly centers evidence in measurement. 

\section{Evidence-Centered Benchmark Design}
\begin{figure*}[!ht]
    \centering
    \includegraphics[trim= 0 10 0 0, clip,scale=0.6]{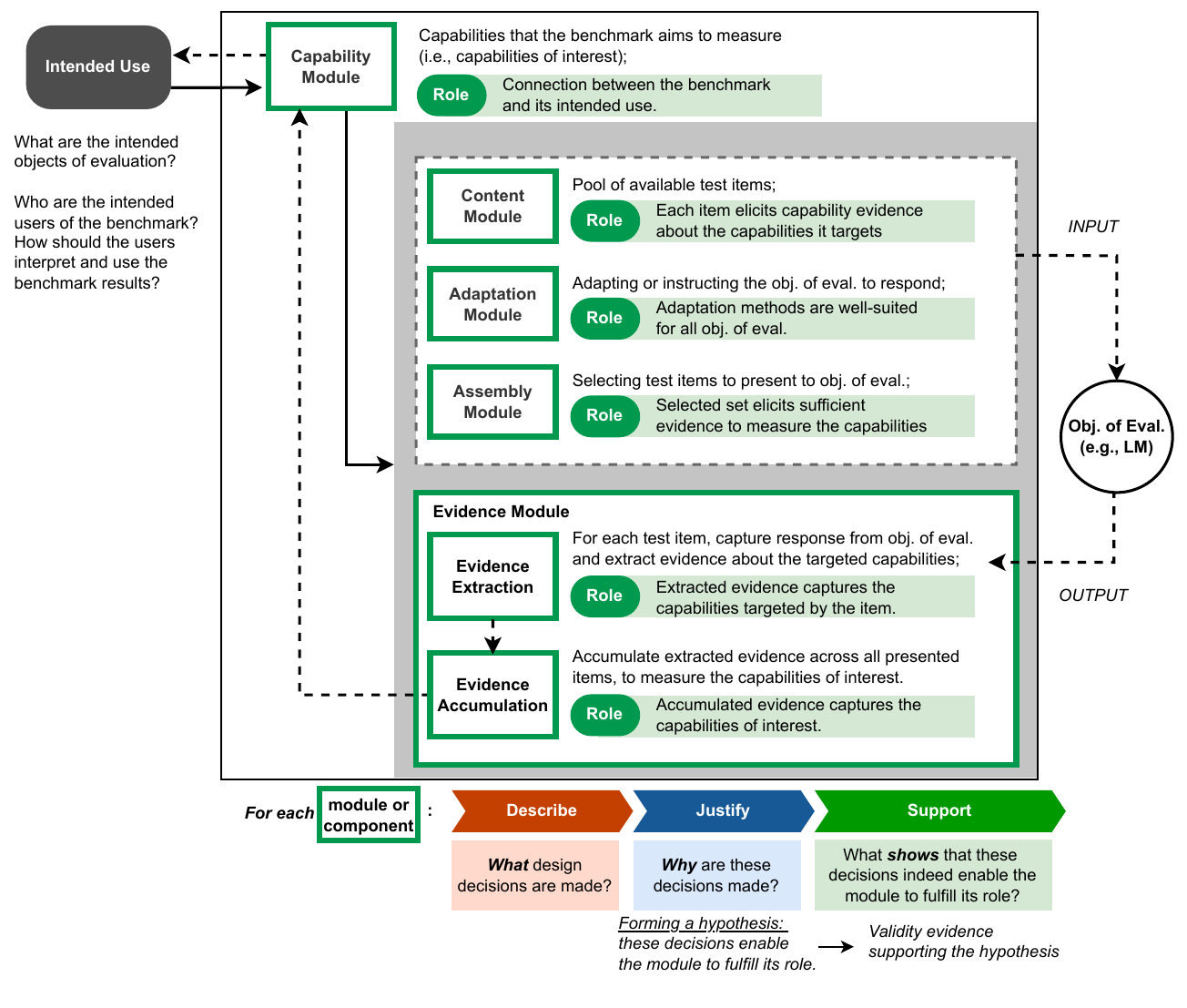}
    \vspace{-8pt}
    \caption{The Evidence-Centered Benchmark Design framework. Solid line arrows indicate the process of designing a benchmark (e.g., designers decide on the intended uses of the benchmark before deciding what capabilities are of interest). The dotted line arrows indicate the process of the benchmark gathering necessary capability evidence.}
    \vspace{-8pt}
    \label{fig:ecbd_large}
\end{figure*}
We consider benchmarking as the process of gathering \emph{capability evidence} from objects of evaluation (e.g., LMs)---i.e., evidence about whether or to what degree those objects have some capabilities of interest. Evidence-Centered Benchmark Design (\framework{}) structures this process into five modules,\footnote{For clarity, in adapting ECD, we have also adapted some of the terminology: i) \textit{module} instead of CAF \textit{model}, as \textit{model} often designates an \textit{NLP model}; ii) \textit{content} instead of \textit{task}, as \textit{task} often refers to a category of problems or challenges that an NLP system aims to solve (e.g., the task of question answering) instead of a test item (e.g., a single exam question).} each of which has a specific role in helping collect the necessary capability evidence: the capability module (\S\ref{sec: capability_module}), the content module (\S\ref{sec: content_module}), the adaptation module (\S\ref{sec: adaptation_module}), the assembly module (\S\ref{sec: assembly_module}), and the evidence module (\S\ref{sec: evidence_module}). 
See Figure~\ref{fig:ecbd_large} for an overview of \framework{}.

For each module, \framework{} breaks down the design process into three required actions. To guide benchmark creation, \framework{} requires benchmark creators to 
i) \textbf{describe} their design choices; 
ii) \textbf{justify} these choices, forming hypotheses about how they ensure that the module accomplishes its role; and 
iii) provide \textbf{support} for these hypotheses, which requires gathering either evidence showing that underlying constructs are well-defined and well-grounded, or \textit{validity evidence}---i.e., evidence about how benchmark design choices might support or threaten the validity of the resulting benchmark measurements. Such evidence can be theoretical (e.g., work conceptualizing a capability) or empirical (e.g., experiments correlating benchmark measurements with some ground truth).  

In addition to helping benchmark creators reflect on their design choices, \framework{} can also support benchmark analysis by benchmark users or third parties by 
drawing attention to whether and how benchmark creators describe and justify their design decisions, and to what extent there is validity evidence supporting these decisions.

We also organize \framework{} in a worksheet of 20 questions to facilitate its use (Appendix~\ref{appendix: worksheet_template}).
Benchmark creators can use this worksheet 
while constructing a benchmark, 
with each question meant to encourage them to reflect on their decisions and make their assumptions explicit. When analyzing existing benchmarks, the worksheet can be completed by referring to available documentation of the benchmark under analysis (e.g., papers, blog posts).

\paragraph*{Benchmark Intended Use}
\label{sec: intended_use}
While establishing the intended use of a benchmark is not a \framework{} module per se, it is a critical step that should precede benchmark design or analysis using \framework{} modules. This first step is important as the validity of the resulting benchmark measurements often concerns whether the benchmark can be used as intended. Articulating a benchmark's intended use requires specifying: 
i) What are the intended objects of evaluation (analogously, ``test takers'')?  
ii) Who are the intended users of the benchmark? and 
iii) How should they interpret and use the benchmark results?\looseness=-1 

A benchmark's design choices should be assessed with respect to its intended use. This intended use might be closely connected to the use context of the evaluated model (e.g., aiming to inform deployment decisions of automatic summarizers for medical records), or it might not be directly connected to any downstream applications (e.g., aiming to compare language models' and humans' linguistic abilities).\looseness=-1 

If the intended use is not clearly stated, benchmark creators risk making choices simply because they are convenient or common practices, likely resulting in a benchmark that does not serve any particular purpose. Furthermore, if the intended use is not clearly communicated to potential users, they could unintentionally misuse the benchmark (e.g., use it to evaluate other objects than those intended), or misinterpret the resulting measurements.

\subsection{Capability Module}
\label{sec: capability_module}
The capability module specifies the capabilities---constructs that the objects of evaluation are thought to exhibit or possess---that the benchmark aims to measure. For NLP models, such capabilities might encompass a wide range of model characteristics and properties, such as reasoning, multilingual understanding, stereotyping, or toxicity.
What capabilities we might want to measure, however, depends on the benchmark's intended use. Thus, this module is also intended to \textit{\textbf{capture the connection between the benchmark and its intended use}}.
To do so, it requires benchmark creators to define the capabilities of interest, justify the aforementioned connection, and provide appropriate grounding for how these capabilities are defined.
This process encourages reflection on the definitions of the capabilities of interest, to ensure that they are i) well-matched to the benchmark's intended use, and ii) well-grounded, as capabilities may be contested and context-dependent (e.g., who are the users of an evaluated model, and what are their needs?).

Theoretical work conceptualizing capabilities, as well as empirical work seeking to understand contexts of use and capabilities relevant to those contexts, could provide evidence that the choices of capabilities and accompanying definitions are well-grounded. For example, \citet{liao2022connecting} conduct survey studies with topical experts and end users to understand what evaluation criteria are important for explainable AI algorithms.

\subsection{Content Module}
\label{sec: content_module}
The content module specifies the pool of available test items that the benchmark could require objects of evaluation to perform or respond to. 
These test items should help \textit{\textbf{elicit evidence about the capabilities of interest}}, so that this evidence can be later extracted from the responses and accumulated to produce measurements of those capabilities (\S\ref{sec: evidence_module}). Note that it is not necessary for each test item to target \textit{all} capabilities of interest, as items can be used in combination (see \S\ref{sec: assembly_module}).

Through the characteristics of the test items, benchmark creators should justify how each test item elicits evidence about the capabilities it targets. Gathering validity evidence for this module---evidence about how the test items help us elicit useful signals about the capabilities of interest---could involve research studies or experts assessing whether test items capture the capabilities of interest.\footnote{In measurement theory, this type of validity evidence is referred to as ``content validity'' \cite{measurement_fairness}.} 
The research study by \citet{blodgett-etal-2021-stereotyping} is such an example, as it examines whether NLP benchmarks meant to measure stereotyping actually measure stereotyping.  
They identify, for instance, test items that contain true facts instead of harmful stereotypes (e.g., ``Afghanistan shares a border with Pakistan. Most people there are Muslim.'' \cite{nangia-etal-2020-crows}). An evaluated model favoring such items is likely not indicative of the model producing harmful stereotypes. Consequently, the prevalence of such test items threatens the validity of these benchmarks.

\subsection{Adaptation Module}
\label{sec: adaptation_module} 
The adaptation module specifies how test conditions are constructed, and how objects of evaluation are instructed (e.g., students) or adapted (e.g., models) for each test item.
For example, benchmarks for evaluating models might employ methods that add examples in few-shot prompting, or adapt models by fine-tuning with examples.
These methods and the data they involve are specified in the adaptation module and should be chosen carefully so as to not confound benchmark measurements. 
The adaptation module should ensure that test conditions and adapted test items are \textit{\textbf{well-suited to all objects of evaluation and not disadvantage some objects}}.\looseness=-1

For example, if a benchmark employs prompting for LMs, some LMs might respond poorly to certain prompt formats \cite{sclar2023quantifying}, thus confounding benchmark results; poor performance might be indicative of this sensitivity to prompt formatting instead of providing meaningful information about the capabilities of interest.

\subsection{Assembly Module}
\label{sec: assembly_module}
The pool of available test items specified by the content module (\S\ref{sec: content_module}) is what the benchmark has available to use. 
The assembly module specifies which items from this pool are actually used by the benchmark for evaluation,
and whether this subset enables practitioners to use the benchmark to \textit{\textbf{gather sufficient evidence for all capabilities of interest}}.\looseness=-1 

The simplest assembly method would be to use all available items. When there are resource constraints (e.g., computational, financial, or time), it may become necessary to consider more sophisticated assembly methods to 
make sure the benchmark measurements remain valid and useful.
For instance, using a smaller set of test items should not introduce an unacceptable amount of measurement error.\looseness=-1 

\subsection{Evidence Module}
\label{sec: evidence_module}
The evidence module specifies how capability evidence is extracted from responses obtained from objects of evaluation (evidence extraction), and how this evidence is accumulated to produce benchmark measurements that capture the capabilities of interest (evidence accumulation). 

\paragraph{Evidence Extraction}
For each presented test item, objects of evaluation produce observable responses (e.g., LM-generated text, token probabilities). Evidence extraction involves specifying \textit{\textbf{what type of responses are elicited from the objects of evaluation and the capability evidence we can infer from these responses}}. 

This process necessarily involves representing the evidence
via some observable variables such as numerical scores (e.g., 1/0 to indicate that a LM-generated text is ``fluent''/``disfluent,'' representing a piece of evidence about the LM's ability to generate fluent text). 
Benchmark creators therefore need to justify and show that these variables actually capture the target capabilities. 
For example, experiments examining the correlation between automatic metrics and human annotations (presumed ground truth) could in some cases provide validity evidence for this component of the module. 

\paragraph{Evidence Accumulation}
Benchmarks involving multiple test items often need to accumulate multiple pieces of extracted evidence to produce measurements of the capabilities of interest to be interpreted and used. This component of the module thus connects observable variables used for evidence extraction to the capability module (Section~\ref{sec: capability_module}): \textbf{\textit{the accumulated evidence should capture the capabilities of interest}}.
For example, the benchmark measurements might be computed as the average of item-level responses, if the distribution of those responses is assumed to follow a normal distribution. Gathering validity evidence could involve testing this assumption about the distribution.\looseness=-1

\section{Case Studies}
To illustrate how our framework guides benchmark analysis and helps foreground possible validity concerns, we apply the \framework{} worksheet to analyze HELM \cite{liang2022holistic}, SuperGLUE \cite{superglue}, and BoolQ \cite{clark-etal-2019-boolq}. 

\subsection{Analyzed Benchmarks}
SuperGLUE aims to be \textit{``a more rigorous test of language understanding''} than its predecessor GLUE \cite{wang-etal-2018-glue}. It includes 8 pre-existing datasets, each corresponding to a \textit{``language understanding task.''} HELM, the most recent benchmark of the three, is meant to be a \textit{``living benchmark''} that is continuously updated. When its accompanying paper was first published, HELM included 15 existing datasets.\footnote{HELM includes two evaluations that seem to be completely independent: a \textit{``core''} evaluation and a supplementary \textit{``targeted''} evaluation. As the main focus of the accompanying paper is on the former, we consider it as a single, independent benchmark that we focus on for our analysis.} BoolQ includes a dataset of real user yes/no queries, which is re-used in both SuperGLUE and HELM.\looseness=-1

These benchmarks are different in many ways: 
they are from different points in time and of various sizes, 
aim to capture different capabilities, and are constructed differently (e.g., BoolQ introduces a novel dataset while SuperGLUE and HELM re-purposes existing datasets).
Due to its flexibility, \framework{} can be applied to all these benchmarks.

\subsection{Method}
The \framework{} worksheet 
for each benchmark was completed by two to three authors of this paper, where one author first read the paper introducing that benchmark, and then re-read it while completing the worksheet. At least two other authors then examined the completed worksheets while reading the paper. We discussed and resolved any ambiguities and inconsistencies that arose during this process both in the phrasing of the worksheet questions and in how to use the information provided in the benchmark papers to answer the questions.  
The completed worksheets can be found at \url{https://github.com/isle-dev/ECBD}.

\subsection{Observations}
We overview key concerns with the design of existing benchmarks that \framework{}'s modules helped us foreground.\looseness=-1

\paragraph{Intended use: Benchmarks' intended uses are vaguely specified.} 
Specifying a benchmark's intended uses is a crucial first step in \framework{}. In examining how the three benchmarks discuss their intended uses, we found little description of who their intended users are and how they should interpret and use the resulting measurements. HELM explicitly states that 
the use and interpretation of benchmark measurements is up to the users to decide for themselves.\footnote{\textit{``[W]e expect the totality of the results we provide are not relevant for every practical use case: we anticipate practitioners should first identify scenarios and metrics pertinent to their use conditions, and then prioritize these scenarios/metrics in interpreting the results of this benchmark.''} \cite[p.~88]{liang2022holistic}\looseness=-1} 
Since validity involves whether benchmark measurements can be used as intended, this lack of information makes the analysis and validation of these benchmarks difficult---in particular, assessing whether measured capabilities are relevant to benchmarks' intended uses.

\paragraph{Capability module: When evaluating complex capabilities, benchmarks seem to break down capabilities of interest into intermediate capabilities that are perhaps easier to measure, but this process is sometimes not explicitly described.}
\begin{figure}[t]
    \centering
    \includegraphics[scale=0.6]{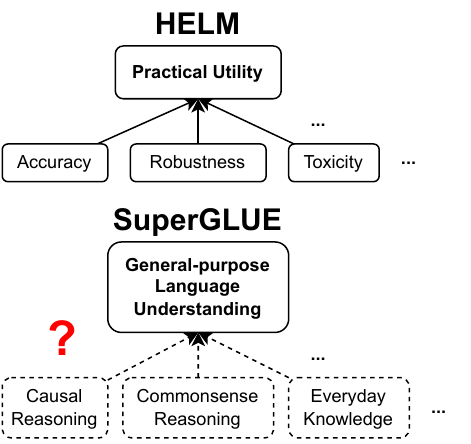}
    \vspace{-6pt}
    \caption{Different levels of capabilities and their connection, in HELM and SuperGLUE. In SuperGLUE, the connection between sub-capabilities (e.g., \textit{``causal reasoning''}) and \textit{``general-purpose language understanding''} is not explained. It is thus denoted by the dotted lines and the question mark.}
    \vspace{-8pt}
    \label{fig:superglue_capability}
\end{figure}
\framework{}'s capability module draws attention towards what capabilities the benchmarks measure and how they are conceptualized. For SuperGLUE, which aims to measure \textit{``general language understanding''} (GLU), we found that the benchmark seems to consider intermediate capabilities of interest that contribute to measuring GLU (see Figure~\ref{fig:superglue_capability}). The paper's descriptions of datasets introduce constructs such as \textit{``causal reasoning,''}\footnote{\textit{``COPA (Choice of Plausible Alternatives, \citet{copa}) is a causal reasoning task [...].''}  \cite{superglue}.} 
treating them as though they are self-evidently relevant to GLU.
However, these additional constructs are not defined, and their connection to GLU is left implied. 
This lack of clarity about capabilities of interest makes it difficult to analyze whether a benchmark properly operationalizes them.
HELM provides more descriptions of choices and definitions of capabilities: it draws an explicit connection between the overall capability under measurement---\textit{``practical utility''}---and the seven intermediate capabilities of interest---e.g., \textit{``accuracy,'' ``calibration''}---which are selected as they are believed to reflect what \textit{``it mean[s] for a system to be useful''} \cite[p.27]{liang2022holistic}. The relationships between these overall and intermediate capabilities, however, might need further interrogation.

\paragraph{Capability module: The capabilities the benchmarks are purportedly measuring are often poorly and/or inconsistently conceptualized.} \framework{} requires benchmark creators not only to say what they want to measure but also to justify why they want to measure it. This helps us foreground inconsistencies in how these capabilities are defined and justified. 
For example, some of the analyzed benchmarks collapse the constructs they are designed to measure with the measurement of those constructs. Specifically, HELM describes e.g., \textit{``accuracy''} (the construct) as an \textit{``umbrella term for the standard accuracy-like metric''}~\cite[p.29]{liang2022holistic} (possible measurements of the construct). This makes it difficult to even know what capability the resulting measurements actually measure.\looseness=-1 

Furthermore, constructs sometimes lack appropriate grounding; for example, HELM conceptualizes constructs like \textit{``fairness,'' ``bias,''} and \textit{``toxicity''} as measurable without requiring `\textit{`knowledge about the broader social context''}~\cite[p.28]{liang2022holistic}. We know from prior work, however, that more often than not, such constructs are contested and depend on the context in which they are applied~\cite{blodgett-etal-2020-language,measurement_fairness}. While such inconsistencies are not necessarily problematic, they can give rise to validity concerns if the benchmark's conceptualizations are not well-justified.\looseness=-1

\paragraph{Content module: For benchmarks re-purposing data, we found little justification connecting the data to capabilities of interest.} 
Pre-existing data that is re-purposed by a benchmark may not be fit for measuring the capabilities that benchmark aims to measure.
By requiring benchmark creators to justify their choice of data, \framework's content module helps highlight potential disconnects between capabilities of interest and the re-purposed data.
For instance, the BoolQ dataset was re-purposed by HELM to measure \textit{``(social) bias,''} among other capabilities. Since this dataset was not designed to elicit evidence about \textit{``bias,''} \framework{} requires HELM to justify and validate the re-use of this data to capture this capability. We found no such justification or validation, which raises doubts about whether the resulting measurement is meaningful.\looseness=-1

\paragraph{Adaptation module: Some benchmarks do not prescribe adaptation methods.} \framework{}'s adaptation module draws attention to the suitability of adaptation methods. Among the benchmarks we inspected, only HELM prescribed an adaptation strategy: few-shot prompting with 5 in-context examples. Once chosen for a given dataset, these examples and the prompt template (e.g., instructions) stay fixed across all test items from that dataset, as well as across all evaluated models. BoolQ and SuperGLUE do not specify whether or how evaluated models need to be adapted. As benchmark users are free to decide for themselves what methods to employ, it might be difficult to meaningfully interpret benchmark measurements when users adopt different adaptation methods for the same benchmark. We recommend that users report what adaptation methods they employ, if such methods are not prescribed by the benchmark.

\paragraph{Assembly module: 
The choices surrounding assembly methods are often taken as self-evidently appropriate.}
We find that the benchmarks we examined do not fully describe or justify the choices of assembly methods. For BoolQ, we only found brief mentions that test items are split into training, development, and test sets, without further elaboration about why and how items are selected to be part of the test set. SuperGLUE uses existing train/dev/test splits from the datasets it re-purposes, but does not describe how these splits were originally constructed or justify the continued use of those splits.
For HELM, a maximum of 1{,}000 test items per dataset are selected for evaluation, but we found no description about the exact selection process. This lack of attention to assembly methods could hinder benchmark creators from considering alternative methods (e.g., selecting test items based on their difficulty \cite{Mishra_Arunkumar_2021}) and reflecting about possible trade-offs different methods might help them make, such as balancing resource constraints with possible threats to validity.\looseness=-1

\paragraph{Evidence module: The way evidence is being extracted and accumulated is often only justified by a desire to follow similar practices as prior work.}
All three benchmarks use automatic metrics to extract evidence, such as exact-match for classification tasks and ROUGE-2 for summarization. These metric scores are then accumulated through functions like F1-score
or by averaging. 
\framework{}'s evidence module requires benchmark creators to justify these choices, particularly with respect to the role they play in extracting and accumulating capability evidence. However, we found that when such choices are justified, the explanations often do not focus on whether or how these methods help capture the capabilities of interest. Instead, benchmark creators only briefly mention that the metrics they use are \textit{``standard''} or \textit{``default''} for a certain task \cite[p.127-137]{liang2022holistic}, or that they are \textit{``follow[ing] prior work''} \cite[p.5-6]{superglue} when choosing metrics or aggregation functions.

We encourage benchmark creators to more carefully consider their choices in the evidence module, including questioning methodology used in prior work, so as not to risk perpetuating the use of currently popular yet potentially unsuitable methodology. Even where methods may be well-justified in prior work, they may not always be well-suited to other contexts (e.g., with differently defined capabilities under measurement), and their appropriateness to new contexts should always be justified.\looseness=-1

\paragraph{Evidence module: Even when new evaluation methods are introduced, we still find little justification for how the methods capture the capabilities of interest.} For example, HELM introduces new automatic metrics to measure \textit{``(social) bias''} through demographic representation. The metric first counts occurrences of words related to each considered demographic group (e.g., \textit{``gomez,'' ``martinez,''} for the group \textit{``Hispanic''}) in model outputs. It then compares the word counts to the uniform distribution (i.e., where every demographic group is equally represented). Some design choices, such as the demographic groups under consideration and their corresponding word lists, are well-described. However, we found little justification for them---for example, why does the benchmark use the demographic groups \textit{``White,'' ``Hispanic,''} and \textit{``Asian''} to measure racial bias? Why is the uniform distribution a suitable reference distribution? Under \framework{}, the creators of HELM would need to justify how these design decisions enable the new metric to capture \textit{``(social) bias.''}

\paragraph{The benchmarks rarely gather validity evidence to support their design choices.} 
All modules in \framework{} require collecting validity evidence to support benchmark design choices. The benchmarks we examined either do not describe collecting such evidence, or acknowledge the need for it but leave gathering it to future work. We encourage benchmark creators to identify and use validity evidence that may already exist, and plan future experiments to gather necessary validity evidence. 
For existing benchmarks, identifying and developing this evidence may require efforts from other researchers, benchmark users, or other practitioners. 
Appropriate incentives from the community could encourage future efforts on gathering validity evidence and on examining how to integrate this evidence into the use of existing benchmarks (e.g., how might a benchmark that includes a metric which is found to be unsuitable be used?). 
For future benchmarks, we strongly encourage benchmark creators to gather and report validity evidence supporting their design choices during the process of benchmark design.

\section{Conclusion}
To guide NLP benchmark creation and analysis, we take inspiration from the evidence-centered design framework from the field of educational testing to propose \framework{} \space(Evidence-Centered Benchmark Design). Our framework formalizes the benchmark design process into five modules that each play a critical role in gathering valid and useful capability evidence---i.e., evidence about whether or to what degree objects of evaluation have some capabilities of interest.

We demonstrated its utility by using it to analyze BoolQ, SuperGLUE, and HELM, finding common practices that threaten the validity of their measurements. For example, for these benchmarks we found more of a focus on describing design choices (e.g., which dataset/metric is used), and less on justifying them and their role in the benchmark. Gathering validity evidence is also rare.\looseness=-1

Future directions include further analyses of our framework's utility in guiding the creation of benchmarks. It is also important to understand how \framework{} helps increase transparency and supports practitioners in achieving a greater understanding of benchmark results and their limitations. Future work might also examine this, for example through user studies with benchmark creators and users.
As \framework{} does not constrain the model inputs and outputs to be textual, we also see it to be applicable or adaptable to multi-modal NLP benchmarks, or to other areas in ML and AI.

\section*{Limitations}
\subsection*{Framework and Worksheet}
While we have developed \framework{} and the accompanying worksheet to guide practitioners through the essential components of the benchmark design process, the questions for each module are not exhaustive, and it is possible that practitioners will identify additional questions particularly relevant to their own benchmark creation or analysis processes.

We also note that assessing a benchmark should involve many criteria beyond validity, such as the provenance of test items (did those who created the data consent to its use?) \cite{rogers-etal-2021-just-think}, privacy (do test items reveal sensitive or personally identifying information?) \cite{huang-etal-2023-privacy}, and the reliability of the measurements (can they be repeated?) \cite{measurement_fairness}.

\subsection*{Case Studies}
The choice of benchmarks to analyze (i.e., BoolQ, SuperGLUE, and HELM) likely also limits our findings. Although there are many differences between them, these benchmarks are unlikely to cover the wide space of possibilities in benchmark design. We have not analyzed, for instance, dynamic benchmarks that create test items instead of relying on existing data \cite{kiela-etal-2021-dynabench}. 

Furthermore, our analysis relied only on the papers introducing each of the three benchmarks, namely the work of \citet{clark-etal-2019-boolq}, of \citet{superglue}, and of \citet{liang2022holistic}. We have not used other sources of information on the benchmarks, such as their official websites and code repositories, which could limit our analysis. On the other hand, only relying on the papers allows us to examine the benchmark creators' reporting practices.

Finally, the case studies are subject to our reading. We could have missed or misinterpreted passages from the analyzed papers.

\section*{Ethical Considerations}
NLP benchmarks not only influence the development and use of specific NLP models---e.g., how performant is a specific model believed to be?---but also help to construct the field's priorities and norms---e.g., what capabilities are researchers developing models towards, and what is seen as evidence of success?
Well-documented and more valid benchmarks run less risk of misguiding benchmark users and stakeholders of evaluated models---potentially avoiding e.g., the costs of optimizing models towards the wrong goal or deploying models with undetected issues, causing harms to users.\looseness=-1 

By proposing a more principled way of designing and analyzing NLP benchmarks, we hope to encourage the construction of well-documented and more valid benchmarks. However, our work could potentially have the unintended, opposite impact of discouraging future work in benchmark design. Although we believe that the benefits of following \framework{} outweigh its costs, extensive documentation in following \framework{}, as well as conducting experiments to gather validity evidence, could be expensive and time-consuming. Finally, although we intend for \framework{} to encourage meaningful reflection during the benchmark design process, as with all documentation there is a risk that it will instead be treated as a checklist.

\section*{Acknowledgements}
We would like to thank Susu Zhang for their helpful discussions and feedback. 
This work is supported by a joint Microsoft Research--Mila grant.
Yu Lu Liu is supported by a Fonds de Recherche du Qu\'{e}bec Nature et Technologies master research scholarship (File \#330991). 
Jackie C.K. Cheung is a consulting researcher at Microsoft Research Montr\'{e}al. 
Finally, we thank the anonymous reviewers for their valuable feedback.
\bibliography{anthology,custom}

\begin{thebibliography}{35}
\expandafter\ifx\csname natexlab\endcsname\relax\def\natexlab#1{#1}\fi

\bibitem[{Aky{\"u}rek et~al.(2022)Aky{\"u}rek, Kocyigit, Paik, and Wijaya}]{akyurek-etal-2022-challenges}
Afra~Feyza Aky{\"u}rek, Muhammed~Yusuf Kocyigit, Sejin Paik, and Derry~Tanti Wijaya. 2022.
\newblock \href {https://doi.org/10.18653/v1/2022.gebnlp-1.9} {Challenges in measuring bias via open-ended language generation}.
\newblock In \emph{Proceedings of the 4th Workshop on Gender Bias in Natural Language Processing (GeBNLP)}, pages 76--76, Seattle, Washington. Association for Computational Linguistics.

\bibitem[{Alzahrani et~al.(2024)Alzahrani, Alyahya, Alnumay, Alrashed, Alsubaie, Almushaykeh, Mirza, Alotaibi, Altwairesh, Alowisheq, Bari, and Khan}]{alzahrani2024when}
Norah Alzahrani, Hisham~Abdullah Alyahya, Yazeed Alnumay, Sultan Alrashed, Shaykhah Alsubaie, Yusef Almushaykeh, Faisal Mirza, Nouf Alotaibi, Nora Altwairesh, Areeb Alowisheq, M~Saiful Bari, and Haidar Khan. 2024.
\newblock When benchmarks are targets: Revealing the sensitivity of large language model leaderboards.
\newblock \emph{arXiv preprint arXiv:2402.01781}.

\bibitem[{American Educational Research~Association(2014)}]{standards}
\&~National Council on Measurement in~Education American Educational Research~Association, American Psychological~Association. 2014.
\newblock \emph{Standards for educational and psychological testing}.
\newblock American Educational Research Association, Lanham, MD.

\bibitem[{Arnold et~al.(2019)Arnold, Bellamy, Hind, Houde, Mehta, Mojsilović, Nair, Ramamurthy, Olteanu, Piorkowski, Reimer, Richards, Tsay, and Varshney}]{arnold2019factsheets}
M.~Arnold, R.~K.~E. Bellamy, M.~Hind, S.~Houde, S.~Mehta, A.~Mojsilović, R.~Nair, K.~Natesan Ramamurthy, A.~Olteanu, D.~Piorkowski, D.~Reimer, J.~Richards, J.~Tsay, and K.~R. Varshney. 2019.
\newblock {FactSheets: Increasing trust in AI services through supplier's declarations of conformity}.
\newblock \emph{IBM Journal of Research and Development}, 63(4/5):6:1--6:13.

\bibitem[{Bandalos(2018)}]{measurement_theory_2018}
Deborah~L. Bandalos. 2018.
\newblock \emph{Measurement theory and applications for the social sciences}.
\newblock The Guilford Press.

\bibitem[{Bender and Friedman(2018)}]{bender-friedman-2018-data}
Emily~M. Bender and Batya Friedman. 2018.
\newblock \href {https://doi.org/10.1162/tacl_a_00041} {Data statements for natural language processing: Toward mitigating system bias and enabling better science}.
\newblock \emph{Transactions of the Association for Computational Linguistics}, 6:587--604.

\bibitem[{Bhardwaj et~al.(2024)Bhardwaj, Gujral, Wu, Zogheib, Maharaj, and Becker}]{bhardwaj2024machine}
Eshta Bhardwaj, Harshit Gujral, Siyi Wu, Ciara Zogheib, Tegan Maharaj, and Christoph Becker. 2024.
\newblock Machine learning data practices through a data curation lens: An evaluation framework.
\newblock \emph{ACM Conference on Fairness, Accountability, and Transparency}.

\bibitem[{Blodgett et~al.(2020)Blodgett, Barocas, Daum{\'e}~III, and Wallach}]{blodgett-etal-2020-language}
Su~Lin Blodgett, Solon Barocas, Hal Daum{\'e}~III, and Hanna Wallach. 2020.
\newblock \href {https://doi.org/10.18653/v1/2020.acl-main.485} {Language (technology) is power: A critical survey of {``}bias{''} in {NLP}}.
\newblock In \emph{Proceedings of the 58th Annual Meeting of the Association for Computational Linguistics}, pages 5454--5476, Online. Association for Computational Linguistics.

\bibitem[{Blodgett et~al.(2021)Blodgett, Lopez, Olteanu, Sim, and Wallach}]{blodgett-etal-2021-stereotyping}
Su~Lin Blodgett, Gilsinia Lopez, Alexandra Olteanu, Robert Sim, and Hanna Wallach. 2021.
\newblock \href {https://doi.org/10.18653/v1/2021.acl-long.81} {Stereotyping {N}orwegian salmon: An inventory of pitfalls in fairness benchmark datasets}.
\newblock In \emph{Proceedings of the 59th Annual Meeting of the Association for Computational Linguistics and the 11th International Joint Conference on Natural Language Processing (Volume 1: Long Papers)}, pages 1004--1015, Online. Association for Computational Linguistics.

\bibitem[{Bowman and Dahl(2021)}]{bowman-dahl-2021-will}
Samuel~R. Bowman and George Dahl. 2021.
\newblock \href {https://doi.org/10.18653/v1/2021.naacl-main.385} {What will it take to fix benchmarking in natural language understanding?}
\newblock In \emph{Proceedings of the 2021 Conference of the North American Chapter of the Association for Computational Linguistics: Human Language Technologies}, pages 4843--4855, Online. Association for Computational Linguistics.

\bibitem[{Clark et~al.(2019)Clark, Lee, Chang, Kwiatkowski, Collins, and Toutanova}]{clark-etal-2019-boolq}
Christopher Clark, Kenton Lee, Ming-Wei Chang, Tom Kwiatkowski, Michael Collins, and Kristina Toutanova. 2019.
\newblock \href {https://doi.org/10.18653/v1/N19-1300} {{B}ool{Q}: Exploring the surprising difficulty of natural yes/no questions}.
\newblock In \emph{Proceedings of the 2019 Conference of the North {A}merican Chapter of the Association for Computational Linguistics: Human Language Technologies, Volume 1 (Long and Short Papers)}, pages 2924--2936, Minneapolis, Minnesota. Association for Computational Linguistics.

\bibitem[{Fleisig et~al.(2023)Fleisig, Amstutz, Atalla, Blodgett, Daum{\'e}~III, Olteanu, Sheng, Vann, and Wallach}]{fleisig-etal-2023-fairprism}
Eve Fleisig, Aubrie Amstutz, Chad Atalla, Su~Lin Blodgett, Hal Daum{\'e}~III, Alexandra Olteanu, Emily Sheng, Dan Vann, and Hanna Wallach. 2023.
\newblock \href {https://doi.org/10.18653/v1/2023.acl-long.343} {{F}air{P}rism: Evaluating fairness-related harms in text generation}.
\newblock In \emph{Proceedings of the 61st Annual Meeting of the Association for Computational Linguistics (Volume 1: Long Papers)}, pages 6231--6251, Toronto, Canada. Association for Computational Linguistics.

\bibitem[{Gebru et~al.(2021)Gebru, Morgenstern, Vecchione, Vaughan, Wallach, III, and Crawford}]{datasheets}
Timnit Gebru, Jamie Morgenstern, Briana Vecchione, Jennifer~Wortman Vaughan, Hanna Wallach, Hal~Daum\'{e} III, and Kate Crawford. 2021.
\newblock \href {https://doi.org/10.1145/3458723} {Datasheets for datasets}.
\newblock \emph{Commun. ACM}, 64(12):86–92.

\bibitem[{Gehrmann et~al.(2021)Gehrmann, Adewumi, Aggarwal, Ammanamanchi, Aremu, Bosselut, Chandu, Clinciu, Das, Dhole, Du, Durmus, Du{\v{s}}ek, Emezue, Gangal, Garbacea, Hashimoto, Hou, Jernite, Jhamtani, Ji, Jolly, Kale, Kumar, Ladhak, Madaan, Maddela, Mahajan, Mahamood, Majumder, Martins, McMillan-Major, Mille, van Miltenburg, Nadeem, Narayan, Nikolaev, Niyongabo~Rubungo, Osei, Parikh, Perez-Beltrachini, Rao, Raunak, Rodriguez, Santhanam, Sedoc, Sellam, Shaikh, Shimorina, Sobrevilla~Cabezudo, Strobelt, Subramani, Xu, Yang, Yerukola, and Zhou}]{GEM_v1}
Sebastian Gehrmann, Tosin Adewumi, Karmanya Aggarwal, Pawan~Sasanka Ammanamanchi, Anuoluwapo Aremu, Antoine Bosselut, Khyathi~Raghavi Chandu, Miruna-Adriana Clinciu, Dipanjan Das, Kaustubh Dhole, Wanyu Du, Esin Durmus, Ond{\v{r}}ej Du{\v{s}}ek, Chris~Chinenye Emezue, Varun Gangal, Cristina Garbacea, Tatsunori Hashimoto, Yufang Hou, Yacine Jernite, Harsh Jhamtani, Yangfeng Ji, Shailza Jolly, Mihir Kale, Dhruv Kumar, Faisal Ladhak, Aman Madaan, Mounica Maddela, Khyati Mahajan, Saad Mahamood, Bodhisattwa~Prasad Majumder, Pedro~Henrique Martins, Angelina McMillan-Major, Simon Mille, Emiel van Miltenburg, Moin Nadeem, Shashi Narayan, Vitaly Nikolaev, Andre Niyongabo~Rubungo, Salomey Osei, Ankur Parikh, Laura Perez-Beltrachini, Niranjan~Ramesh Rao, Vikas Raunak, Juan~Diego Rodriguez, Sashank Santhanam, Jo{\~a}o Sedoc, Thibault Sellam, Samira Shaikh, Anastasia Shimorina, Marco~Antonio Sobrevilla~Cabezudo, Hendrik Strobelt, Nishant Subramani, Wei Xu, Diyi Yang, Akhila Yerukola, and Jiawei Zhou. 2021.
\newblock \href {https://doi.org/10.18653/v1/2021.gem-1.10} {The {GEM} benchmark: Natural language generation, its evaluation and metrics}.
\newblock In \emph{Proceedings of the 1st Workshop on Natural Language Generation, Evaluation, and Metrics (GEM 2021)}, pages 96--120, Online. Association for Computational Linguistics.

\bibitem[{Gehrmann et~al.(2022)Gehrmann, Bhattacharjee, Mahendiran, Wang, Papangelis, Madaan, McMillan-Major, Shvets, Upadhyay, Yao, Wilie, Bhagavatula, You, Thomson, Garbacea, Wang, Deutsch, Xiong, Jin, Gkatzia, Radev, Clark, Durmus, Ladhak, Ginter, Winata, Strobelt, Hayashi, Novikova, Kanerva, Chim, Zhou, Clive, Maynez, Sedoc, Juraska, Dhole, Chandu, Perez-Beltrachini, Ribeiro, Tunstall, Zhang, Pushkarna, Creutz, White, Kale, Eddine, Daheim, Subramani, Dusek, Liang, Ammanamanchi, Zhu, Puduppully, Kriz, Shahriyar, Cardenas, Mahamood, Osei, Cahyawijaya, Štajner, Montella, Shailza, Jolly, Mille, Hasan, Shen, Adewumi, Raunak, Raheja, Nikolaev, Tsai, Jernite, Xu, Sang, Liu, and Hou}]{GEM_v2}
Sebastian Gehrmann, Abhik Bhattacharjee, Abinaya Mahendiran, Alex Wang, Alexandros Papangelis, Aman Madaan, Angelina McMillan-Major, Anna Shvets, Ashish Upadhyay, Bingsheng Yao, Bryan Wilie, Chandra Bhagavatula, Chaobin You, Craig Thomson, Cristina Garbacea, Dakuo Wang, Daniel Deutsch, Deyi Xiong, Di~Jin, Dimitra Gkatzia, Dragomir Radev, Elizabeth Clark, Esin Durmus, Faisal Ladhak, Filip Ginter, Genta~Indra Winata, Hendrik Strobelt, Hiroaki Hayashi, Jekaterina Novikova, Jenna Kanerva, Jenny Chim, Jiawei Zhou, Jordan Clive, Joshua Maynez, João Sedoc, Juraj Juraska, Kaustubh Dhole, Khyathi~Raghavi Chandu, Laura Perez-Beltrachini, Leonardo F.~R. Ribeiro, Lewis Tunstall, Li~Zhang, Mahima Pushkarna, Mathias Creutz, Michael White, Mihir~Sanjay Kale, Moussa~Kamal Eddine, Nico Daheim, Nishant Subramani, Ondrej Dusek, Paul~Pu Liang, Pawan~Sasanka Ammanamanchi, Qi~Zhu, Ratish Puduppully, Reno Kriz, Rifat Shahriyar, Ronald Cardenas, Saad Mahamood, Salomey Osei, Samuel Cahyawijaya, Sanja Štajner, Sebastien Montella,
  Shailza, Shailza Jolly, Simon Mille, Tahmid Hasan, Tianhao Shen, Tosin Adewumi, Vikas Raunak, Vipul Raheja, Vitaly Nikolaev, Vivian Tsai, Yacine Jernite, Ying Xu, Yisi Sang, Yixin Liu, and Yufang Hou. 2022.
\newblock \href {http://arxiv.org/abs/2206.11249} {Gemv2: Multilingual nlg benchmarking in a single line of code}.

\bibitem[{Goldfarb-Tarrant et~al.(2023)Goldfarb-Tarrant, Ungless, Balkir, and Blodgett}]{goldfarb-tarrant-etal-2023-prompt}
Seraphina Goldfarb-Tarrant, Eddie Ungless, Esma Balkir, and Su~Lin Blodgett. 2023.
\newblock \href {https://doi.org/10.18653/v1/2023.findings-acl.139} {This prompt is measuring {\textless}mask{\textgreater}: evaluating bias evaluation in language models}.
\newblock In \emph{Findings of the Association for Computational Linguistics: ACL 2023}, pages 2209--2225, Toronto, Canada. Association for Computational Linguistics.

\bibitem[{Gururangan et~al.(2018)Gururangan, Swayamdipta, Levy, Schwartz, Bowman, and Smith}]{gururangan-etal-2018-annotation}
Suchin Gururangan, Swabha Swayamdipta, Omer Levy, Roy Schwartz, Samuel Bowman, and Noah~A. Smith. 2018.
\newblock \href {https://doi.org/10.18653/v1/N18-2017} {Annotation artifacts in natural language inference data}.
\newblock In \emph{Proceedings of the 2018 Conference of the North {A}merican Chapter of the Association for Computational Linguistics: Human Language Technologies, Volume 2 (Short Papers)}, pages 107--112, New Orleans, Louisiana. Association for Computational Linguistics.

\bibitem[{Huang et~al.(2023)Huang, Gupta, Zhong, Li, and Chen}]{huang-etal-2023-privacy}
Yangsibo Huang, Samyak Gupta, Zexuan Zhong, Kai Li, and Danqi Chen. 2023.
\newblock \href {https://doi.org/10.18653/v1/2023.emnlp-main.921} {Privacy implications of retrieval-based language models}.
\newblock In \emph{Proceedings of the 2023 Conference on Empirical Methods in Natural Language Processing}, pages 14887--14902, Singapore. Association for Computational Linguistics.

\bibitem[{Jacobs and Wallach(2021)}]{measurement_fairness}
Abigail~Z. Jacobs and Hanna Wallach. 2021.
\newblock \href {https://doi.org/10.1145/3442188.3445901} {Measurement and fairness}.
\newblock In \emph{Proceedings of the 2021 ACM Conference on Fairness, Accountability, and Transparency}, FAccT '21, page 375–385, New York, NY, USA. Association for Computing Machinery.

\bibitem[{Kiela et~al.(2021)Kiela, Bartolo, Nie, Kaushik, Geiger, Wu, Vidgen, Prasad, Singh, Ringshia, Ma, Thrush, Riedel, Waseem, Stenetorp, Jia, Bansal, Potts, and Williams}]{kiela-etal-2021-dynabench}
Douwe Kiela, Max Bartolo, Yixin Nie, Divyansh Kaushik, Atticus Geiger, Zhengxuan Wu, Bertie Vidgen, Grusha Prasad, Amanpreet Singh, Pratik Ringshia, Zhiyi Ma, Tristan Thrush, Sebastian Riedel, Zeerak Waseem, Pontus Stenetorp, Robin Jia, Mohit Bansal, Christopher Potts, and Adina Williams. 2021.
\newblock \href {https://doi.org/10.18653/v1/2021.naacl-main.324} {Dynabench: Rethinking benchmarking in {NLP}}.
\newblock In \emph{Proceedings of the 2021 Conference of the North American Chapter of the Association for Computational Linguistics: Human Language Technologies}, pages 4110--4124, Online. Association for Computational Linguistics.

\bibitem[{Liang et~al.(2022)Liang, Bommasani, Lee, Tsipras, Soylu, Yasunaga, Zhang, Narayanan, Wu, Kumar et~al.}]{liang2022holistic}
Percy Liang, Rishi Bommasani, Tony Lee, Dimitris Tsipras, Dilara Soylu, Michihiro Yasunaga, Yian Zhang, Deepak Narayanan, Yuhuai Wu, Ananya Kumar, et~al. 2022.
\newblock Holistic evaluation of language models.
\newblock \emph{arXiv preprint arXiv:2211.09110}.

\bibitem[{Liao and Xiao(2023)}]{liao2023rethinking}
Q.~Vera Liao and Ziang Xiao. 2023.
\newblock Rethinking model evaluation as narrowing the socio-technical gap.
\newblock \emph{arXiv preprint arXiv:2306.03100}.

\bibitem[{Liao et~al.(2022)Liao, Zhang, Luss, Doshi-Velez, and Dhurandhar}]{liao2022connecting}
Q~Vera Liao, Yunfeng Zhang, Ronny Luss, Finale Doshi-Velez, and Amit Dhurandhar. 2022.
\newblock {Connecting Algorithmic Research and Usage Contexts: A Perspective of Contextualized Evaluation for Explainable AI}.
\newblock In \emph{Proceedings of the AAAI Conference on Human Computation and Crowdsourcing}, volume~10, pages 147--159.

\bibitem[{Mishra and Arunkumar(2021)}]{Mishra_Arunkumar_2021}
Swaroop Mishra and Anjana Arunkumar. 2021.
\newblock \href {https://doi.org/10.1609/aaai.v35i15.17599} {How robust are model rankings : A leaderboard customization approach for equitable evaluation}.
\newblock \emph{Proceedings of the AAAI Conference on Artificial Intelligence}, 35(15):13561--13569.

\bibitem[{Mislevy et~al.(2003)Mislevy, Steinberg, and Almond}]{mislevy_structure_ECD}
Robert~J Mislevy, Linda~S Steinberg, and Russell~G Almond. 2003.
\newblock Focus article: On the structure of educational assessments.
\newblock \emph{Measurement: Interdisciplinary research and perspectives}, 1(1):3--62.

\bibitem[{Mitchell et~al.(2019)Mitchell, Wu, Zaldivar, Barnes, Vasserman, Hutchinson, Spitzer, Raji, and Gebru}]{mitchell2019model}
Margaret Mitchell, Simone Wu, Andrew Zaldivar, Parker Barnes, Lucy Vasserman, Ben Hutchinson, Elena Spitzer, Inioluwa~Deborah Raji, and Timnit Gebru. 2019.
\newblock Model cards for model reporting.
\newblock In \emph{Proceedings of the Conference on Fairness, Accountability, and Transparency}, pages 220--229.

\bibitem[{Nangia et~al.(2020)Nangia, Vania, Bhalerao, and Bowman}]{nangia-etal-2020-crows}
Nikita Nangia, Clara Vania, Rasika Bhalerao, and Samuel~R. Bowman. 2020.
\newblock \href {https://doi.org/10.18653/v1/2020.emnlp-main.154} {{C}row{S}-pairs: A challenge dataset for measuring social biases in masked language models}.
\newblock In \emph{Proceedings of the 2020 Conference on Empirical Methods in Natural Language Processing (EMNLP)}, pages 1953--1967, Online. Association for Computational Linguistics.

\bibitem[{Roemmele et~al.(2011)Roemmele, Bejan, and Gordon}]{copa}
Melissa Roemmele, Cosmin~Adrian Bejan, and Andrew~S Gordon. 2011.
\newblock Choice of plausible alternatives: An evaluation of commonsense causal reasoning.
\newblock In \emph{2011 AAAI Spring Symposium Series}.

\bibitem[{Rogers et~al.(2021)Rogers, Baldwin, and Leins}]{rogers-etal-2021-just-think}
Anna Rogers, Timothy Baldwin, and Kobi Leins. 2021.
\newblock \href {https://doi.org/10.18653/v1/2021.findings-emnlp.414} {{`}just what do you think you{'}re doing, dave?{'} a checklist for responsible data use in {NLP}}.
\newblock In \emph{Findings of the Association for Computational Linguistics: EMNLP 2021}, pages 4821--4833, Punta Cana, Dominican Republic. Association for Computational Linguistics.

\bibitem[{Sclar et~al.(2023)Sclar, Choi, Tsvetkov, and Suhr}]{sclar2023quantifying}
Melanie Sclar, Yejin Choi, Yulia Tsvetkov, and Alane Suhr. 2023.
\newblock \href {http://arxiv.org/abs/2310.11324} {Quantifying language models' sensitivity to spurious features in prompt design or: How i learned to start worrying about prompt formatting}.

\bibitem[{Srivastava et~al.(2022)Srivastava, Rastogi, Rao, Shoeb, Abid, Fisch, Brown, Santoro, Gupta, Garriga-Alonso, Kluska, Lewkowycz, Agarwal, Power, Ray, Warstadt, Kocurek, Safaya, Tazarv, Xiang, Parrish, Nie, Hussain, Askell, Dsouza, Slone, Rahane, Iyer, Andreassen, Madotto, Santilli, Stuhlmüller, Dai, La, Lampinen, Zou, Jiang, Chen, Vuong, Gupta, Gottardi, Norelli, Venkatesh, Gholamidavoodi, Tabassum, Menezes, Kirubarajan, Mullokandov, Sabharwal, Herrick, Efrat, Erdem, Karakaş, Roberts, Loe, Zoph, Bojanowski, Özyurt, Hedayatnia, Neyshabur, Inden, Stein, Ekmekci, Lin, Howald, Diao, Dour, Stinson, Argueta, Ramírez, Singh, Rathkopf, Meng, Baral, Wu, Callison-Burch, Waites, Voigt, Manning, Potts, Ramirez, Rivera, Siro, Raffel, Ashcraft, Garbacea, Sileo, Garrette, Hendrycks, Kilman, Roth, Freeman, Khashabi, Levy, González, Perszyk, Hernandez, Chen, Ippolito, Gilboa, Dohan, Drakard, Jurgens, Datta, Ganguli, Emelin, Kleyko, Yuret, Chen, Tam, Hupkes, Misra, Buzan, Mollo, Yang, Lee, Shutova, Cubuk, Segal,
  Hagerman, Barnes, Donoway, Pavlick, Rodola, Lam, Chu, Tang, Erdem, Chang, Chi, Dyer, Jerzak, Kim, Manyasi, Zheltonozhskii, Xia, Siar, Martínez-Plumed, Happé, Chollet, Rong, Mishra, Winata, de~Melo, Kruszewski, Parascandolo, Mariani, Wang, Jaimovitch-López, Betz, Gur-Ari, Galijasevic, Kim, Rashkin, Hajishirzi, Mehta, Bogar, Shevlin, Schütze, Yakura, Zhang, Wong, Ng, Noble, Jumelet, Geissinger, Kernion, Hilton, Lee, Fisac, Simon, Koppel, Zheng, Zou, Kocoń, Thompson, Kaplan, Radom, Sohl-Dickstein, Phang, Wei, Yosinski, Novikova, Bosscher, Marsh, Kim, Taal, Engel, Alabi, Xu, Song, Tang, Waweru, Burden, Miller, Balis, Berant, Frohberg, Rozen, Hernandez-Orallo, Boudeman, Jones, Tenenbaum, Rule, Chua, Kanclerz, Livescu, Krauth, Gopalakrishnan, Ignatyeva, Markert, Dhole, Gimpel, Omondi, Mathewson, Chiafullo, Shkaruta, Shridhar, McDonell, Richardson, Reynolds, Gao, Zhang, Dugan, Qin, Contreras-Ochando, Morency, Moschella, Lam, Noble, Schmidt, He, Colón, Metz, Şenel, Bosma, Sap, ter Hoeve, Farooqi, Faruqui,
  Mazeika, Baturan, Marelli, Maru, Quintana, Tolkiehn, Giulianelli, Lewis, Potthast, Leavitt, Hagen, Schubert, Baitemirova, Arnaud, McElrath, Yee, Cohen, Gu, Ivanitskiy, Starritt, Strube, Swędrowski, Bevilacqua, Yasunaga, Kale, Cain, Xu, Suzgun, Tiwari, Bansal, Aminnaseri, Geva, Gheini, T, Peng, Chi, Lee, Krakover, Cameron, Roberts, Doiron, Nangia, Deckers, Muennighoff, Keskar, Iyer, Constant, Fiedel, Wen, Zhang, Agha, Elbaghdadi, Levy, Evans, Casares, Doshi, Fung, Liang, Vicol, Alipoormolabashi, Liao, Liang, Chang, Eckersley, Htut, Hwang, Miłkowski, Patil, Pezeshkpour, Oli, Mei, Lyu, Chen, Banjade, Rudolph, Gabriel, Habacker, Delgado, Millière, Garg, Barnes, Saurous, Arakawa, Raymaekers, Frank, Sikand, Novak, Sitelew, LeBras, Liu, Jacobs, Zhang, Salakhutdinov, Chi, Lee, Stovall, Teehan, Yang, Singh, Mohammad, Anand, Dillavou, Shleifer, Wiseman, Gruetter, Bowman, Schoenholz, Han, Kwatra, Rous, Ghazarian, Ghosh, Casey, Bischoff, Gehrmann, Schuster, Sadeghi, Hamdan, Zhou, Srivastava, Shi, Singh, Asaadi, Gu,
  Pachchigar, Toshniwal, Upadhyay, Shyamolima, Debnath, Shakeri, Thormeyer, Melzi, Reddy, Makini, Lee, Torene, Hatwar, Dehaene, Divic, Ermon, Biderman, Lin, Prasad, Piantadosi, Shieber, Misherghi, Kiritchenko, Mishra, Linzen, Schuster, Li, Yu, Ali, Hashimoto, Wu, Desbordes, Rothschild, Phan, Wang, Nkinyili, Schick, Kornev, Telleen-Lawton, Tunduny, Gerstenberg, Chang, Neeraj, Khot, Shultz, Shaham, Misra, Demberg, Nyamai, Raunak, Ramasesh, Prabhu, Padmakumar, Srikumar, Fedus, Saunders, Zhang, Vossen, Ren, Tong, Zhao, Wu, Shen, Yaghoobzadeh, Lakretz, Song, Bahri, Choi, Yang, Hao, Chen, Belinkov, Hou, Hou, Bai, Seid, Zhao, Wang, Wang, Wang, and Wu}]{srivastava2022imitation}
Aarohi Srivastava, Abhinav Rastogi, Abhishek Rao, Abu Awal~Md Shoeb, Abubakar Abid, Adam Fisch, Adam~R. Brown, Adam Santoro, Aditya Gupta, Adrià Garriga-Alonso, Agnieszka Kluska, Aitor Lewkowycz, Akshat Agarwal, Alethea Power, Alex Ray, Alex Warstadt, Alexander~W. Kocurek, Ali Safaya, Ali Tazarv, Alice Xiang, Alicia Parrish, Allen Nie, Aman Hussain, Amanda Askell, Amanda Dsouza, Ambrose Slone, Ameet Rahane, Anantharaman~S. Iyer, Anders Andreassen, Andrea Madotto, Andrea Santilli, Andreas Stuhlmüller, Andrew Dai, Andrew La, Andrew Lampinen, Andy Zou, Angela Jiang, Angelica Chen, Anh Vuong, Animesh Gupta, Anna Gottardi, Antonio Norelli, Anu Venkatesh, Arash Gholamidavoodi, Arfa Tabassum, Arul Menezes, Arun Kirubarajan, Asher Mullokandov, Ashish Sabharwal, Austin Herrick, Avia Efrat, Aykut Erdem, Ayla Karakaş, B.~Ryan Roberts, Bao~Sheng Loe, Barret Zoph, Bartłomiej Bojanowski, Batuhan Özyurt, Behnam Hedayatnia, Behnam Neyshabur, Benjamin Inden, Benno Stein, Berk Ekmekci, Bill~Yuchen Lin, Blake Howald,
  Cameron Diao, Cameron Dour, Catherine Stinson, Cedrick Argueta, César~Ferri Ramírez, Chandan Singh, Charles Rathkopf, Chenlin Meng, Chitta Baral, Chiyu Wu, Chris Callison-Burch, Chris Waites, Christian Voigt, Christopher~D. Manning, Christopher Potts, Cindy Ramirez, Clara~E. Rivera, Clemencia Siro, Colin Raffel, Courtney Ashcraft, Cristina Garbacea, Damien Sileo, Dan Garrette, Dan Hendrycks, Dan Kilman, Dan Roth, Daniel Freeman, Daniel Khashabi, Daniel Levy, Daniel~Moseguí González, Danielle Perszyk, Danny Hernandez, Danqi Chen, Daphne Ippolito, Dar Gilboa, David Dohan, David Drakard, David Jurgens, Debajyoti Datta, Deep Ganguli, Denis Emelin, Denis Kleyko, Deniz Yuret, Derek Chen, Derek Tam, Dieuwke Hupkes, Diganta Misra, Dilyar Buzan, Dimitri~Coelho Mollo, Diyi Yang, Dong-Ho Lee, Ekaterina Shutova, Ekin~Dogus Cubuk, Elad Segal, Eleanor Hagerman, Elizabeth Barnes, Elizabeth Donoway, Ellie Pavlick, Emanuele Rodola, Emma Lam, Eric Chu, Eric Tang, Erkut Erdem, Ernie Chang, Ethan~A. Chi, Ethan Dyer, Ethan
  Jerzak, Ethan Kim, Eunice~Engefu Manyasi, Evgenii Zheltonozhskii, Fanyue Xia, Fatemeh Siar, Fernando Martínez-Plumed, Francesca Happé, Francois Chollet, Frieda Rong, Gaurav Mishra, Genta~Indra Winata, Gerard de~Melo, Germán Kruszewski, Giambattista Parascandolo, Giorgio Mariani, Gloria Wang, Gonzalo Jaimovitch-López, Gregor Betz, Guy Gur-Ari, Hana Galijasevic, Hannah Kim, Hannah Rashkin, Hannaneh Hajishirzi, Harsh Mehta, Hayden Bogar, Henry Shevlin, Hinrich Schütze, Hiromu Yakura, Hongming Zhang, Hugh~Mee Wong, Ian Ng, Isaac Noble, Jaap Jumelet, Jack Geissinger, Jackson Kernion, Jacob Hilton, Jaehoon Lee, Jaime~Fernández Fisac, James~B. Simon, James Koppel, James Zheng, James Zou, Jan Kocoń, Jana Thompson, Jared Kaplan, Jarema Radom, Jascha Sohl-Dickstein, Jason Phang, Jason Wei, Jason Yosinski, Jekaterina Novikova, Jelle Bosscher, Jennifer Marsh, Jeremy Kim, Jeroen Taal, Jesse Engel, Jesujoba Alabi, Jiacheng Xu, Jiaming Song, Jillian Tang, Joan Waweru, John Burden, John Miller, John~U. Balis,
  Jonathan Berant, Jörg Frohberg, Jos Rozen, Jose Hernandez-Orallo, Joseph Boudeman, Joseph Jones, Joshua~B. Tenenbaum, Joshua~S. Rule, Joyce Chua, Kamil Kanclerz, Karen Livescu, Karl Krauth, Karthik Gopalakrishnan, Katerina Ignatyeva, Katja Markert, Kaustubh~D. Dhole, Kevin Gimpel, Kevin Omondi, Kory Mathewson, Kristen Chiafullo, Ksenia Shkaruta, Kumar Shridhar, Kyle McDonell, Kyle Richardson, Laria Reynolds, Leo Gao, Li~Zhang, Liam Dugan, Lianhui Qin, Lidia Contreras-Ochando, Louis-Philippe Morency, Luca Moschella, Lucas Lam, Lucy Noble, Ludwig Schmidt, Luheng He, Luis~Oliveros Colón, Luke Metz, Lütfi~Kerem Şenel, Maarten Bosma, Maarten Sap, Maartje ter Hoeve, Maheen Farooqi, Manaal Faruqui, Mantas Mazeika, Marco Baturan, Marco Marelli, Marco Maru, Maria Jose~Ramírez Quintana, Marie Tolkiehn, Mario Giulianelli, Martha Lewis, Martin Potthast, Matthew~L. Leavitt, Matthias Hagen, Mátyás Schubert, Medina~Orduna Baitemirova, Melody Arnaud, Melvin McElrath, Michael~A. Yee, Michael Cohen, Michael Gu,
  Michael Ivanitskiy, Michael Starritt, Michael Strube, Michał Swędrowski, Michele Bevilacqua, Michihiro Yasunaga, Mihir Kale, Mike Cain, Mimee Xu, Mirac Suzgun, Mo~Tiwari, Mohit Bansal, Moin Aminnaseri, Mor Geva, Mozhdeh Gheini, Mukund~Varma T, Nanyun Peng, Nathan Chi, Nayeon Lee, Neta Gur-Ari Krakover, Nicholas Cameron, Nicholas Roberts, Nick Doiron, Nikita Nangia, Niklas Deckers, Niklas Muennighoff, Nitish~Shirish Keskar, Niveditha~S. Iyer, Noah Constant, Noah Fiedel, Nuan Wen, Oliver Zhang, Omar Agha, Omar Elbaghdadi, Omer Levy, Owain Evans, Pablo Antonio~Moreno Casares, Parth Doshi, Pascale Fung, Paul~Pu Liang, Paul Vicol, Pegah Alipoormolabashi, Peiyuan Liao, Percy Liang, Peter Chang, Peter Eckersley, Phu~Mon Htut, Pinyu Hwang, Piotr Miłkowski, Piyush Patil, Pouya Pezeshkpour, Priti Oli, Qiaozhu Mei, Qing Lyu, Qinlang Chen, Rabin Banjade, Rachel~Etta Rudolph, Raefer Gabriel, Rahel Habacker, Ramón~Risco Delgado, Raphaël Millière, Rhythm Garg, Richard Barnes, Rif~A. Saurous, Riku Arakawa, Robbe
  Raymaekers, Robert Frank, Rohan Sikand, Roman Novak, Roman Sitelew, Ronan LeBras, Rosanne Liu, Rowan Jacobs, Rui Zhang, Ruslan Salakhutdinov, Ryan Chi, Ryan Lee, Ryan Stovall, Ryan Teehan, Rylan Yang, Sahib Singh, Saif~M. Mohammad, Sajant Anand, Sam Dillavou, Sam Shleifer, Sam Wiseman, Samuel Gruetter, Samuel~R. Bowman, Samuel~S. Schoenholz, Sanghyun Han, Sanjeev Kwatra, Sarah~A. Rous, Sarik Ghazarian, Sayan Ghosh, Sean Casey, Sebastian Bischoff, Sebastian Gehrmann, Sebastian Schuster, Sepideh Sadeghi, Shadi Hamdan, Sharon Zhou, Shashank Srivastava, Sherry Shi, Shikhar Singh, Shima Asaadi, Shixiang~Shane Gu, Shubh Pachchigar, Shubham Toshniwal, Shyam Upadhyay, Shyamolima, Debnath, Siamak Shakeri, Simon Thormeyer, Simone Melzi, Siva Reddy, Sneha~Priscilla Makini, Soo-Hwan Lee, Spencer Torene, Sriharsha Hatwar, Stanislas Dehaene, Stefan Divic, Stefano Ermon, Stella Biderman, Stephanie Lin, Stephen Prasad, Steven~T. Piantadosi, Stuart~M. Shieber, Summer Misherghi, Svetlana Kiritchenko, Swaroop Mishra, Tal
  Linzen, Tal Schuster, Tao Li, Tao Yu, Tariq Ali, Tatsu Hashimoto, Te-Lin Wu, Théo Desbordes, Theodore Rothschild, Thomas Phan, Tianle Wang, Tiberius Nkinyili, Timo Schick, Timofei Kornev, Timothy Telleen-Lawton, Titus Tunduny, Tobias Gerstenberg, Trenton Chang, Trishala Neeraj, Tushar Khot, Tyler Shultz, Uri Shaham, Vedant Misra, Vera Demberg, Victoria Nyamai, Vikas Raunak, Vinay Ramasesh, Vinay~Uday Prabhu, Vishakh Padmakumar, Vivek Srikumar, William Fedus, William Saunders, William Zhang, Wout Vossen, Xiang Ren, Xiaoyu Tong, Xinran Zhao, Xinyi Wu, Xudong Shen, Yadollah Yaghoobzadeh, Yair Lakretz, Yangqiu Song, Yasaman Bahri, Yejin Choi, Yichi Yang, Yiding Hao, Yifu Chen, Yonatan Belinkov, Yu~Hou, Yufang Hou, Yuntao Bai, Zachary Seid, Zhuoye Zhao, Zijian Wang, Zijie~J. Wang, Zirui Wang, and Ziyi Wu. 2022.
\newblock \href {http://arxiv.org/abs/2206.04615} {Beyond the imitation game: Quantifying and extrapolating the capabilities of language models}.

\bibitem[{Wagstaff(2012)}]{ML_that_matters}
Kiri~L. Wagstaff. 2012.
\newblock Machine learning that matters.
\newblock In \emph{Proceedings of the 29th International Coference on International Conference on Machine Learning}, ICML'12, page 1851–1856, Madison, WI, USA. Omnipress.

\bibitem[{Wang et~al.(2019)Wang, Pruksachatkun, Nangia, Singh, Michael, Hill, Levy, and Bowman}]{superglue}
Alex Wang, Yada Pruksachatkun, Nikita Nangia, Amanpreet Singh, Julian Michael, Felix Hill, Omer Levy, and Samuel Bowman. 2019.
\newblock {SuperGLUE: A Stickier Benchmark for General-Purpose Language Understanding Systems}.
\newblock \emph{Advances in neural information processing systems}, 32.

\bibitem[{Wang et~al.(2018)Wang, Singh, Michael, Hill, Levy, and Bowman}]{wang-etal-2018-glue}
Alex Wang, Amanpreet Singh, Julian Michael, Felix Hill, Omer Levy, and Samuel Bowman. 2018.
\newblock \href {https://doi.org/10.18653/v1/W18-5446} {{GLUE}: A multi-task benchmark and analysis platform for natural language understanding}.
\newblock In \emph{Proceedings of the 2018 {EMNLP} Workshop {B}lackbox{NLP}: Analyzing and Interpreting Neural Networks for {NLP}}, pages 353--355, Brussels, Belgium. Association for Computational Linguistics.

\bibitem[{Xiao et~al.(2023)Xiao, Zhang, Lai, and Liao}]{xiao-etal-2023-evaluating-evaluation}
Ziang Xiao, Susu Zhang, Vivian Lai, and Q.~Vera Liao. 2023.
\newblock \href {https://doi.org/10.18653/v1/2023.emnlp-main.676} {Evaluating evaluation metrics: A framework for analyzing {NLG} evaluation metrics using measurement theory}.
\newblock In \emph{Proceedings of the 2023 Conference on Empirical Methods in Natural Language Processing}, pages 10967--10982, Singapore. Association for Computational Linguistics.

\end{thebibliography}
\bibliographystyle{acl_natbib}

\appendix

\section{Worksheet Template}
\label{appendix: worksheet_template}
\subsection*{Introduction}
Evidence-Centered Benchmark Design (\framework{}) is a framework that formalizes the benchmark design process. It requires first specifying the \textbf{intended use} of the benchmark (including specifying the objects of evaluation). The process is then broken down into five modules:
\begin{compactenum}[i)]
    \item \textbf{Capability module}: capabilities that the benchmark aims to measure.
    \item \textbf{Content module}: pool of test items that draw out responses from the objects.
    \item \textbf{Adaptation module}: adapting or instructing the objects to complete the tasks.
    \item \textbf{Assembly module}: selecting from the pool of test items to build the set used for evaluation.
    \item \textbf{Evidence module}: extracting and accumulating evidence about the capabilities of interest from responses produced by the objects. 
\end{compactenum}

\paragraph{}This worksheet provides guidance on how to create a new benchmark or analyze an existing benchmark following \framework{}. It can be completed from different perspectives: as the creator of a new benchmark, as the custodian or the user of an existing benchmark, or as a third-party analyzing benchmarks, etc. Each module contains three questions:
\begin{compactenum}[-]
    \item \textbf{Describe}: What design decisions did the benchmark creators make for this module?
    \item \textbf{Justify}: Why did the benchmark creators make these decisions? This involves forming a hypothesis that the decisions allow the module to accomplish its role in the process of gathering necessary capability evidence.
    \item \textbf{Support}: What validity evidence do the benchmark creators have to support the above hypothesis? In other words, what shows that the module indeed accomplishes its role?
\end{compactenum}

\paragraph{} This worksheet is not a checklist, and it is not required to answer each question perfectly. These questions are meant to encourage reflection and validation of benchmark design decisions, as well as to guide benchmark documentation.

\subsection*{Benchmark Name and Reference(s)}
The references are the source of information used to complete this worksheet. For example, a third-party analyzing an existing benchmark may choose to use the academic publication introducing said benchmark as their source of information. Other sources of information could be blog posts, official websites, or code repositories accompanying the benchmark. 

\noindent[ANSWER HERE]

\subsection*{Who is filing the worksheet?}
From what perspective is this worksheet completed? In other words, what is the relation between the person(s) completing this worksheet and the benchmark that is the focus of this worksheet? 

\noindent[ANSWER HERE]

\subsection{Intended Use}
\textbf{Q1 - Who/What are the intended objects of evaluation?} Elaboration on the objects of evaluation (e.g., their assumed capabilities, demographic information for human objects of evaluation, etc.) helps us better understand whether the benchmark is suitable for all intended objects of evaluation.

\noindent[ANSWER HERE]

\paragraph{Q2 - What is the intended use of the benchmark? Who are the intended users of the benchmark?} Benchmark results aim to provide insights about the objects of evaluation: how are users meant to use these insights? 

\noindent[ANSWER HERE]

\subsection{Capability Module}
The capability module specifies the capabilities that the benchmark aims to evaluate. The term “capability” refers to a construct (e.g., quality criteria, skill, etc.) that the objects of evaluation are thought to exhibit or possess. Capabilities often cannot be directly observed or directly measured, thus requiring the benchmark to indirectly measure them by gathering necessary evidence about said capabilities. \\

\noindent \textbf{Q3 - DESCRIBE: i) What are the capabilities of interest? ii) How is each one defined, and under what context is each one defined?}

\noindent[ANSWER HERE]

\paragraph{} Additional recommended questions to consider so to further clarify and contextualize the definitions (in benchmark analysis: as presented by the benchmark):
\begin{itemize}
    \item How does the definition used by the benchmark differ from other existing definitions of this capability? \newline [ANSWER HERE]
    \item How does this capability differ from other similarly defined capabilities?\newline [ANSWER HERE]
\end{itemize}

\paragraph{Q4 - JUSTIFY: How are the capabilities of interest connected to the intended use of the benchmark (specified in Q2)? Are the capabilities theoretically attainable by the objects to be evaluated?} Explain the interest in measuring the capabilities in Q3 and question whether it may be impossible for the objects of evaluation to have said capabilities.

\noindent[ANSWER HERE]\\

\noindent \textbf{Q5 - SUPPORT: What validity evidence do the benchmark creators offer to support the choice and definition of capabilities of interest?} 

\noindent[ANSWER HERE]

\subsection{Content Module}
The content module specifies test items that the benchmark could require objects of evaluation to perform or to respond to. The test items should elicit evidence about some capability of interest, so that said capability evidence can be later extracted from the responses and aggregated to produce a measurement of said capability.

\paragraph{Q6 - DESCRIBE:} \textbf{i) Characterize the test items}. Most often, NLP evaluation relies on input data, so this step could involve describing the data that is available to the benchmark to use, how the data is obtained, etc. \textbf{ii)  Which capabilities of interest does each test item aim to capture?} Each item can aim to capture one or several capabilities amongst those listed in Q3.

\noindent[ANSWER HERE]\\

\noindent \textbf{Q7 - JUSTIFY: How does each test item elicit evidence about its target capabilities? Justify via the characteristics of the test items (Q6).} 

\noindent[ANSWER HERE]

\paragraph{Q8 - SUPPORT: What evidence do the benchmark creators offer to support content validity of the test items?} In other words, we question whether the test items captures capabilities of interest. Content validity is often based on analysis by external experts or benchmark users.

\noindent[ANSWER HERE]

\subsection{Adaptation Module}
When evaluating humans, the benchmark might instruct them to perform a task by providing instructions, training exercises, demonstrations, etc. When evaluating models/systems, there are also myriad methods that i) modify the models/systems (e.g., fine-tuning), or ii) format or add onto the input (e.g., adding examples in few-shot prompting). These adaptation methods should be chosen carefully so as to not confound evaluation results.\\

\noindent \textbf{Q9 - DESCRIBE: Given an input, how are the objects of evaluation adapted or instructed to provide the output?}

\noindent[ANSWER HERE]\\

\noindent \textbf{Q10 - JUSTIFY: Elaborate on the suitability of the adaptation methods for all intended objects of evaluation.}

\noindent[ANSWER HERE]\\

\noindent \textbf{Q11 - SUPPORT: What validity evidence do benchmark designers offer that supports the choice of the adaptation methods?}

\noindent[ANSWER HERE]

\subsection{Assembly Module}
Test items specified by the content module are what the benchmark could use. The assembly module concerns what items from that pool will actually be used by the benchmark for evaluation, and whether this set allows the benchmark to gather sufficient evidence.\\

\noindent \textbf{Q12 - DESCRIBE: How many test items are chosen to assemble the subset used for evaluation? What factors inform this selection?}

\noindent[ANSWER HERE]\\

\noindent \textbf{Q13 - JUSTIFY: How does the described assembly method ensure that the produced subset elicits sufficient evidence for all capabilities of interest?  }

\noindent[ANSWER HERE]\\

\noindent \textbf{Q14 - SUPPORT: What validity evidence do the benchmark creators offer to support the choice of assembly methods?}

\noindent[ANSWER HERE]

\subsection{Evidence Module}
\subsubsection{Evidence Extraction Component}
In response to each presented test item, objects of evaluation produce observable behaviors (referred to as “responses”) which are captured by the benchmark. From these responses, the benchmark extracts evidence about capabilities of interest that said test item targets (referred to as ``salient evidence''). 

\paragraph{Q15 - DESCRIBE: For each test item,} \textbf{i) What responses are captured and used for evidence extraction?} When evaluating humans, many types of responses can be captured: selection in multiple-choice questions, long-form answers, response time, etc. Similarly, the benchmark can use the generated text (decoded in a certain way), token probabilities, running time, etc. \textbf{ii) How is evidence extracted and represented?} 

\noindent[ANSWER HERE]\\

\noindent \textbf{Q16 - JUSTIFY: How does the extracted evidence capture the capabilities of interest? }

\noindent[ANSWER HERE]\\

\noindent \textbf{Q17 - SUPPORT: What validity evidence do the benchmark creators offer to support the choice of evidence extraction method?}

\noindent[ANSWER HERE]\\

\subsubsection{Evidence Accumulation Component}

\noindent \textbf{Q18 - DESCRIBE: How is the evidence accumulated to draw insights about the objects of evaluation in terms of capabilities of interest?}

\noindent[ANSWER HERE]\\

\noindent \textbf{Q19 - JUSTIFY: How does the method of accumulating evidence capture capabilities of interest? }

\noindent[ANSWER HERE]\\

\noindent \textbf{Q20 - SUPPORT: What validity evidence do the benchmark creators offer to support the choice of evidence accumulation method?}

\noindent[ANSWER HERE]\\
\section{Glossary}
We compile terminology used in the present paper and in the \framework{} worksheet in Table~\ref{tab_glossary}.
\begin{table*}
  \centering
  \begin{tabular}{|L{2.5cm}|L{12cm}|}
    \hline
    \textbf{Term} & \textbf{Meaning} \\ \hline
    \textit{Objects of evaluation} & Models, systems, people, etc. that are to be evaluated. \\ \hline
    \textit{Capability} & Quality criteria, ability, skill, etc. that characterizes the objects of evaluation. They are very often not observable nor directly measurable. \\ \hline
    \textit{Capability evidence} & Evidence indicating whether or to what degree an object of evaluation has the capability of interest. For example, a language model (object of evaluation) detecting the grammatical error in “their going to the mall.” can be a piece of evidence supporting the belief that the model has grammatical knowledge (capability of interest). \\ \hline
    \textit{Benchmarking (verb); a benchmark (noun)} & We view benchmarking as a process of gathering capability evidence from the objects of evaluation about the capabilities of interest. A benchmark is a collection of measurement instruments that supports the above process. \\ \hline
    \textit{Benchmark results} & The final product of benchmarking, often in the form of numerical scores (e.g., ratio), rankings, or categorization (e.g., detecting that an object of evaluation is “biased”). The results inform benchmark users about the objects of evaluation, about to whether or to what degree the object has the capabilities of interest. \\ \hline
    \textit{Validity Evidence} & Evidence supporting whether the benchmark results can be interpreted as it is originally intended to be interpreted, whether the benchmark can be used as it is originally intended to be used. In other words, it is evidence supporting that the capability evidence gathered is actually meaningful with respect to the intended uses of the benchmark. Validity evidence can be theoretical or empirical/ \\ \hline
    \textit{Validity; validation} & Validity is the degree to which all the accumulated validity evidence supports the intended interpretation of benchmark results for the intended use of the benchmark. Validation is thus the process of accumulating validity evidence \\ \hline
    \textit{Test item} & A single evaluation instance of the benchmark that objects of evaluation can be asked to perform or respond to in order to obtain outputs or behaviours from them. \\ \hline
    \textit{Response} & Outputs or behaviours from the objects of evaluation in response to a test item presented to them. These are expected to be observable. For example, a matrix of token probabilities can be a response from a language model. The decoded text that the model generated can also be a response. What response to capture is a benchmark design decision.\\ \hline
    \textit{Context} (in the capability module) & Where and how the objects of evaluation are intended to be used or intended to operate under. Context can involve the types of model/system users, other stakeholders, the domain of application, the linguistic phenomena the systems are meant to represent, etc. The definition of capabilities can greatly vary depending on context (e.g., informativeness of some texts varies for expert vs. non-expert readers) \\ \hline
  \end{tabular}
  \newline\newline
  \caption{Glossary}\label{tab_glossary}
\end{table*}
\end{document}